\documentclass[a4paper]{article}

\usepackage{amssymb}
\usepackage{amsmath}
\usepackage{graphicx}
\usepackage{subcaption}
\newcommand{\X}{\ensuremath{\mathbf{X}}}

\renewcommand{\H}{\ensuremath{\mathbf{H}}}

\renewcommand{\r}{\ensuremath{\mathbf{r}}}
\newcommand{\x}{\ensuremath{\mathbf{x}}}
\newcommand{\y}{\ensuremath{\mathbf{y}}}
\newcommand{\h}{\ensuremath{\mathbf{h}}}
\renewcommand{\b}{\ensuremath{\boldsymbol{\beta}}}
\newcommand{\s}{\ensuremath{\boldsymbol{\sigma}}}

\begin{document}

\title{Per-sample Prediction Intervals for Extreme Learning Machines}

\author{Anton Akusok$^1$, Yoan Miche$^2$, Kaj-Mikael Bj\"{o}rk$^3$ \and Amaury Lendasse$^4$}
\date{%
    $^1$Arcada University of Applied Sciences, Helsinki, Finland\\
    $^2$Nokia Solutions and Networks Group, Espoo, Finland\\
    $^3$Risklab at Arcada UAS, Helsinki, Finland\\
    $^4$Department of Mechanical and Industrial Engineering and\\ The Iowa Informatics Initiative, The University of Iowa, Iowa City, USA
}

\maketitle

\begin{abstract}

Prediction intervals in supervised Machine Learning bound the region where the true outputs of new samples may fall. They are necessary in the task of separating reliable predictions of a trained model from near random guesses, minimizing the rate of False Positives, and other problem-specific tasks in applied Machine Learning. Many real problems have heteroscedastic stochastic outputs, which explains the need of input-dependent prediction intervals.

This paper proposes to estimate the input-dependent prediction intervals by a separate Extreme Learning Machine model, using variance of its predictions as a correction term accounting for the model uncertainty. The variance is estimated from the model's linear output layer with a weighted Jackknife method. The methodology is very fast, robust to heteroscedastic outputs, and handles both extremely large datasets and insufficient amount of training data.

\end{abstract}

\section{Introduction}

Practical applications of machine learning can be problematic in the sense that developers and practitioneers often do not fully trust in their own predictions. A fundamental reason for this mistrust can be found in the fact that Mean Squared Error (MSE) and other error measures averaged over a dataset are commonly used to evaluate performance of a method or compare different methods. Averaged error measures are unfit for business processes where each particular sample is important, as it represents a customer or other existing entity~\cite{akusok_twostage_2014}. On the other hand, applied Machine Learning models might skip some data samples, because they are only a part of a bigger process structure, and uncertain data might be given to human experts to be handled~\cite{j._hegedus_methodology_2011}.

The trust problem can be solved by computing a sample-specific confidence value~\cite{pevec_input_2014}. Then predictions with high confidence (and enough trust in them) are used, while data samples with uncertain predictions are passed to the next analytical stage. The Machine Learning model works as a filter, solving “easy cases” automatically with confident predictions, and reducing the amount of data remaining to be analyzed~\cite{akusok_arbitrary_2015}.

Let $\{\x_i, y_i\}, \ i \in [1,N]$ be a dataset where outputs $y$ are independently drawn from a normal distribution conditioned on inputs $\x$:
\begin{equation}
	y = \mathcal{N}(f(\x), \ \sigma^2(\x)) = f(\x) + \mathcal{N}(0, \ \sigma^2(\x))
	\label{eq:1}
\end{equation}
This dataset has heteroscedastic noise because the variance is not constant. A common homoscedasticity assumption simplifies formula~\eqref{eq:1} to $y = f(\x) + N(0, \ \sigma^2)$ but removes the ability to separate confident predictions from uncertain ones. 

The heteroscedastisity of outputs is a reasonable assumption because applied Machine Learning problems often have stochastic outputs. Such outputs do not have a single correct value for the given input. The variance of random noise in outputs may be assumed equal because the noise is independent of the inputs, but the same assumption cannot be made about the variance of the stochastic outputs because they certainly depend on the inputs. 

This work focuses on prediction intervals specifically for Extreme Learning Machines (ELM)~\cite{huang_extreme_2004,lendasse_advances_2016}. ELM is a fast non-linear model with universal approximation ability~\cite{huang_universal_2006}. It has a feed-forward neural network structure but with randomly fixed hidden layer weights, so only the linear output layer needs to be trained. With a large hidden layer and L2-regularization~\cite{tikhonov_solution_1963} ELM exhibit stable predictions~\cite{miche_tropelm_2011}, that are not affected by a particular initialization of the random hidden layer weights. It is an excelled Machine Learning tool to solve applied problems~\cite{akusok_mdelm_2015,termenon_brain_2016} with simple formulation, little to no hyper-parameters, performance at the state-of-the-art level~\cite{huang_local_2015,sovilj_extreme_2016,z._huang_efficient_2017} and scalable to Big Data~\cite{akusok_highperformance_2015,swaney_efficient_2015}.

The idea of the method is to use an ELM to predict an output $f(\x)$, and a second ELM to estimate its conditional variance $\sigma^2(\x) = (y - f(\x))^2$. Furthermore, a variance analysis is done on the predictions of the second ELM. It provides upper and lower boundaries for the predicted variance. These boundaries describe the model uncertainty for samples with little similar training data available, and make the methodology uniformly applicable to different problems.

The rest of the paper is organized as following. The following section describes state-of-the-art in prediction intervals estimation, and how the proposed solution differs from the rest. Section 2 describes Extreme Learning Machines and the proposed methodology. Section 3 analyses the method performance on small artificial and real world datasets. Section 4 presents the results on huge real world dataset, and describes the runtime requirements compared to the original ELM. Section 5 summarizes the findings.

\subsection{State-of-the-Art}

Prediction with uncertainty in a well-known task. Probabilistic methods can obviously formulate a solution. Prediction intervals are available in Bayesian formulation of ELM~\cite{e._soria-olivas_belm:_2011,chen_variational_2016}, including per-sample PI~\cite{shang_confidence-weighted_2015} though the applicability is limited due to the quadratic computational cost in the number of data samples.

Fuzzy nonlinear regression~\cite{he_fuzzy_2016} approach exists for problems having fuzzy inputs or outputs. It applies random weights neural networks with non-iterative training similar to ELM, but formulates the solution in terms of fuzzy sets theory~\cite{asai_linear_1982}. Such a native fuzzy approach allows for a detailed investigation of the effects of uncertainty on learning of a method~\cite{wang_study_2015,wang_noniterative_2017}, and has important practical applications~\cite{ashfaq_fuzziness_2017} for fuzzy data problems.

Without runtime limitation, good results are achieved with model independent methods~\cite{pevec_prediction_2015} based on clustering of input data and re-sampling. Clustering of inputs and repetitive model re-training during the re-sampling both scale poorly with data size, and would limit the performance of ELM otherwise capable of processing billions of data samples~\cite{akusok_highperformance_2015}.

A specific case~\cite{Lin2017} of model-independent approach limited to linear models (with arbitrary solution algorithm and hyper-parameters) provides good results for heteroscedastic datasets (\cite{Lin2017}, \textit{supplementary materials}), and suits for ELM output layer solution as well. The method applies to any amount of training data, and will benefit from huge datasets by producing more independent models in its ensemble part. Unfortunately, it does not output prediction intervals directly.

The scope of this paper is constrained to fast ways of computing prediction intervals of outputs, tailored specifically for Extreme Learning Machine. The proposed solution works especially well in conjunction with ELM, re-using some heavy computational parts as shown in the next section. A fast runtime is one of the the key features of ELM, making it valuable for practical applications and Big Data processing. Another key feature of ELM is approximation of complex unknown functions, and the proposed method approximates prediction intervals of model outputs in similar fashion without probabilistic or fuzzy set notations.

\section{Methodology}
\label{sec:method}

This section starts by introducing the Extreme Learning Machine. It continues with the prediction intervals idea, and its implementation suitable for ELM. The section concludes with a formal description of an algorithm.

\subsection{Extreme Learning Machine}

The Extreme Learning Machine~\cite{huang_extreme_2006} model is formulated as a feed-forward neural network with a single hidden layer. It has $d$ input and $L$ hidden neurons. Solution is given for one output neuron; in case of many output neurons each one has an independent solution. The hidden layer weights $\mathbf{W}_{d \times L}$ are initialized with random noise and fixed. Often an extra input neuron with the constant $+1$ value is added to function as bias.

Hidden layer neurons apply a non-linear transformation function $\phi(\cdot)$ to their output. Typical functions are sigmoid or hyperbolic tangent, but this function may be omitted to add linear neurons. For $N$ input data samples gathered in a matrix $\mathbf X_{N \times d}$, the hidden layer output matrix $\mathbf H_{N \times L}$ is:
\begin{equation}
    \mathbf H_{i,j} = \phi(\sum_{k=1}^d \mathbf X_{i,k} \mathbf W_{k,j}), \ i \in [1, N], \ j \in [1,L]
\end{equation}
where the function $\phi()$ is applied element-wise. In matrix notation, the formula simplifies to $\mathbf H = \phi(\mathbf X \mathbf W)$.

The output layer of ELM is a linear regression problem $\mathbf H \b = \mathbf y$, that is over-determined in real cases with more data samples than hidden neurons ($N > L$). The output weights $\b$ are given by an ordinary least squares solution $\b = \H^\dagger \y$ computed with the Moore-Penrose pseudoinverse~\cite{rao_generalized_1972} $\H^\dagger$ of matrix $\H$.

Random initialization may decrease the performance of a naive ELM. This problem is completely solved by including L2 regularization in the output layer solution. The linear regression problem becomes:
\begin{equation}
     (\mathbf H^T \mathbf H + \gamma \mathbf I)\beta = (\mathbf H^T \mathbf y) \label{eq:ELM1}
\end{equation}
where $\gamma$ is L2-regularization parameter optimized by validation. With L2 regularization and a large number of hidden neurons, ELM performance becomes stable and unaffected by a particular random initialization of $\mathbf W$~\cite{huang_extreme_2012}.

\subsection{Prediction Intervals}

Assume a stochastic output $y$ with i.i.d. distribution conditioned on the inputs $\mathbf x$ as in equation~\eqref{eq:1}. Model prediction $\hat{y} = \hat{f}(\mathbf x)$ estimates only the mean value of an output, and ignores its stochastic nature.

Prediction intervals (PI) offer a simple way of describing the uncertainty of the output $y$ by estimating the boundaries on its value, such that the true output $y$ lies between those boundaries with the given probability $\alpha$. For normally distributed outputs~\eqref{eq:1} the prediction intervals at the confidence level $\alpha$ can be modelled by
\begin{equation}
    \text{PI}(\x) = \hat{f}(\x) \pm \Phi^{-1}(\alpha)\sigma(\x),
    \label{eq:4}
\end{equation}
where $\Phi^{-1}()$ is an inverse cumulative distribution function, i.e. $\Phi^{-1}(95\%) \approx 1.96$.

The maximum likelihood estimator for the variance $\sigma^2$ of a homoscedastic output $y$  is given by Mean Squared Error~\cite{bishop_pattern_2006}. However, it provides uniform prediction intervals that fit poorly to practical applications of Machine Learning. 

An estimation of variance in linear regression is a well-researched topic, with plethora of theoretical~\cite{shao_heteroscedasticity-robustness_1987} and experimental~\cite{pevec_prediction_2015} results available. Variance of heteroscedastic model predictions $\hat{y}$ can be computed with the Bienaym\'e formula~\cite{loeve_probability_1955,johnson_applied_2002} from the variance of model weights $\b$. However, variance of the predicted outputs corresponds to confidence intervals and does not describe the range of possible true outputs $y$.

\begin{figure*}[p]
    \centering
    \begin{subfigure}[t]{0.49\textwidth}
        \centering
        \includegraphics[width=\textwidth]{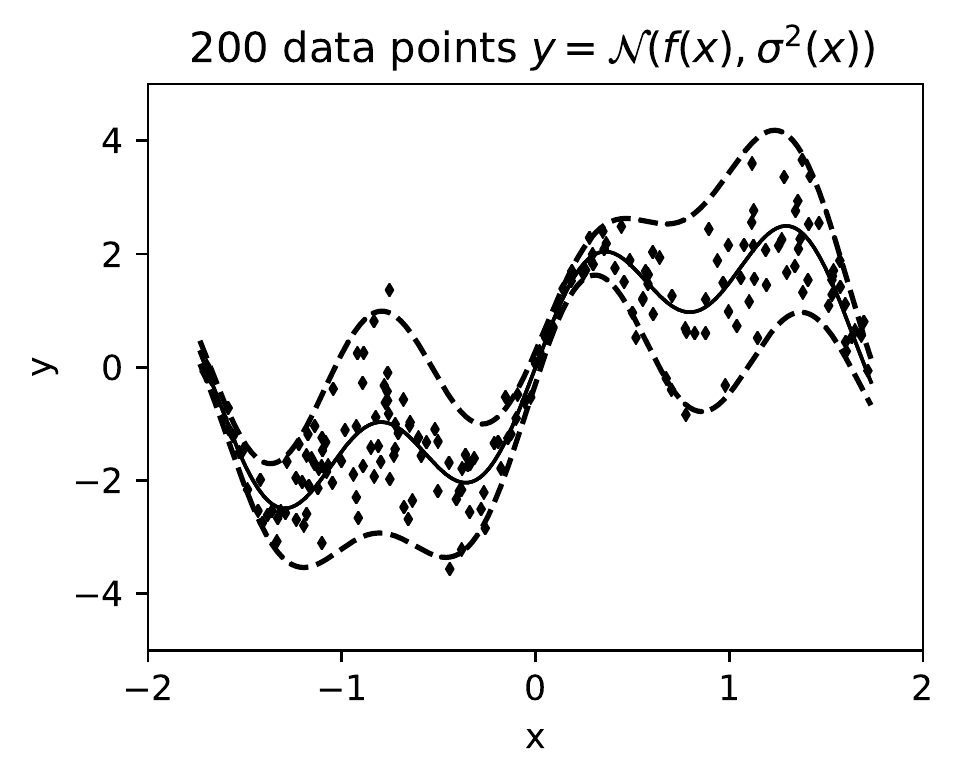}
        \caption{Training data points, true function $f(x)$ and $2\sigma(x)$ boundaries.}
    \end{subfigure}%
    ~ 
    \begin{subfigure}[t]{0.49\textwidth}
        \centering
        \includegraphics[width=\textwidth]{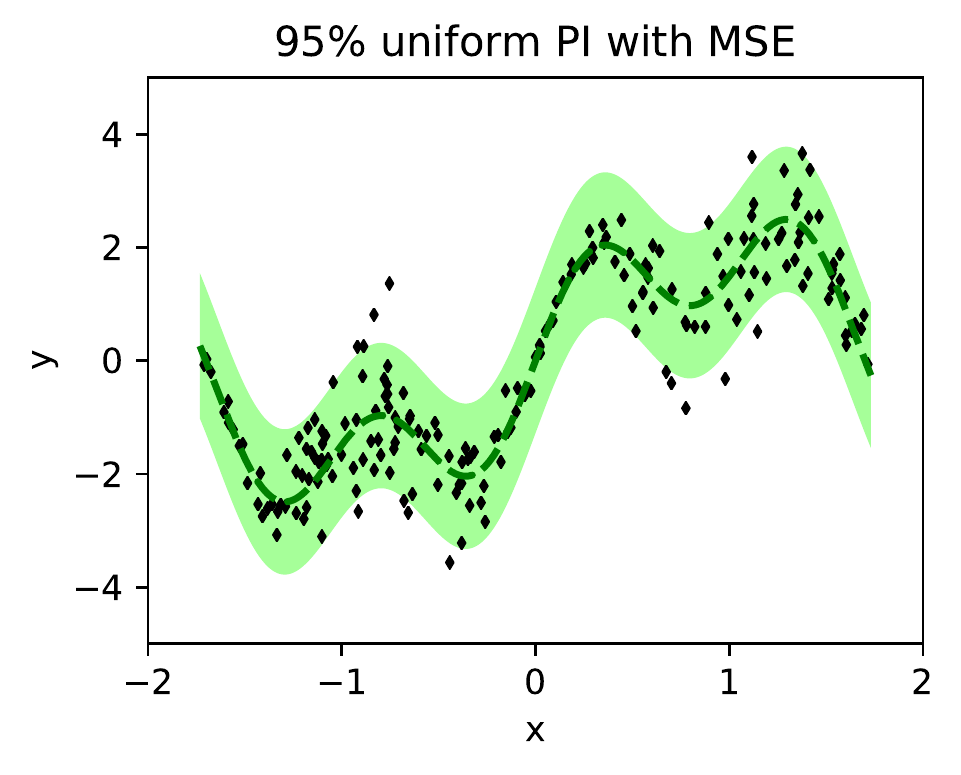}
        \caption{Prediction intervals with MSE that estimate uniform $f(x) \pm 2\sigma$ boundaries.}
    \end{subfigure}
        \begin{subfigure}[t]{0.49\textwidth}
        \centering
        \includegraphics[width=\textwidth]{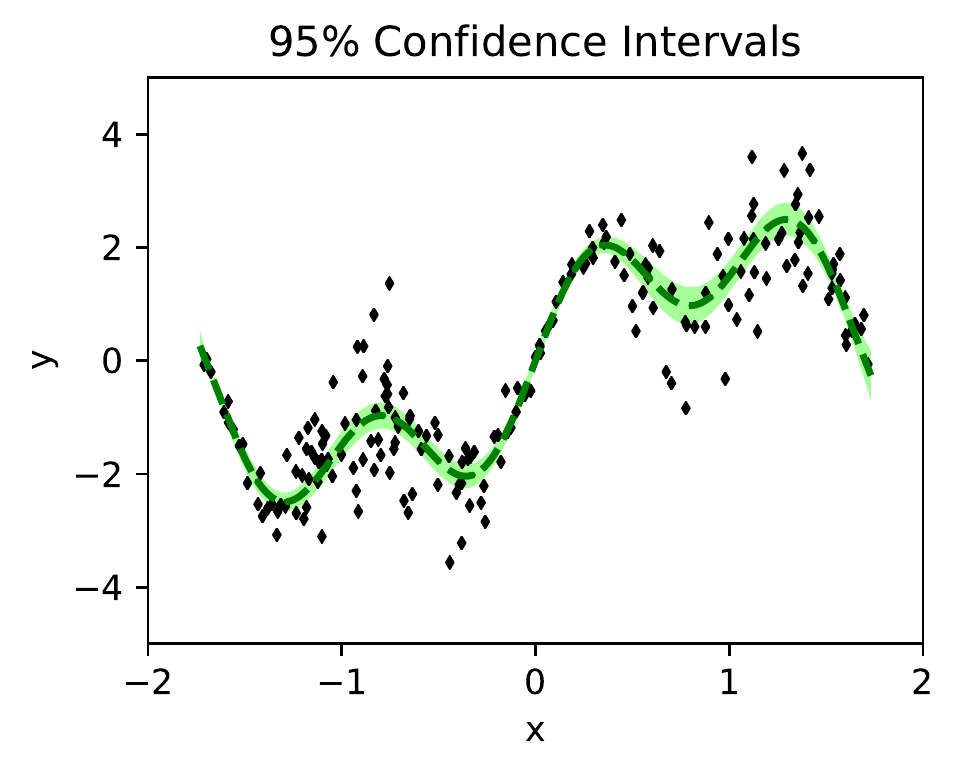}
        \caption{Confidence intervals of model predictions $\hat{y}$.}
    \end{subfigure}%
    ~ 
    \begin{subfigure}[t]{0.49\textwidth}
        \centering
        \includegraphics[width=\textwidth]{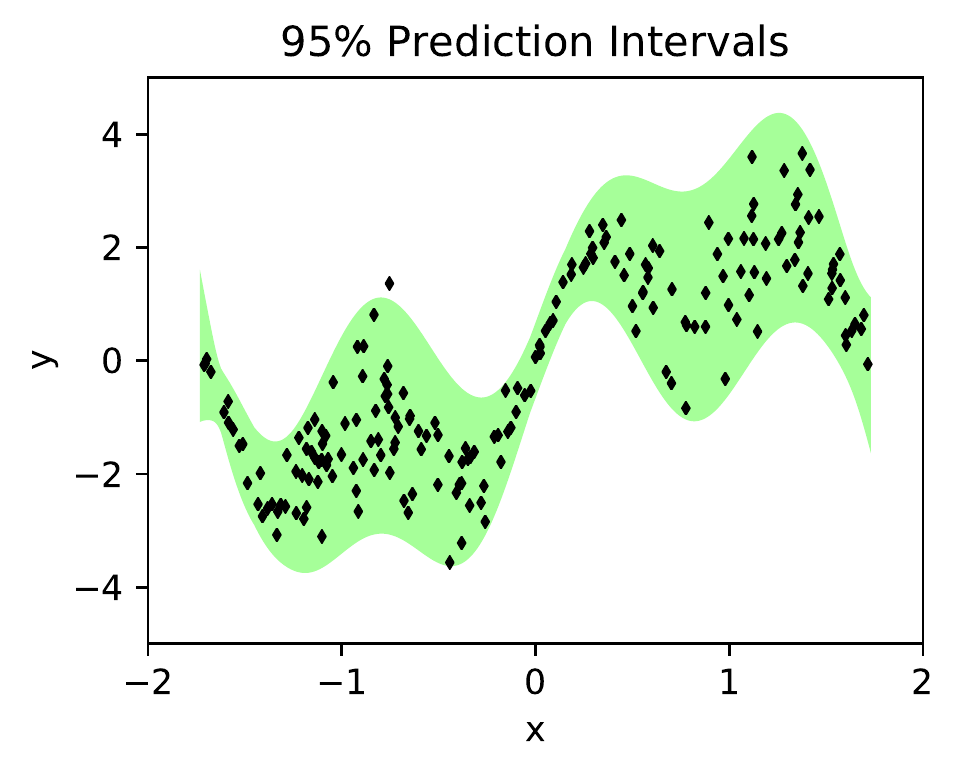}
        \caption{Heteroscedastic prediction intervals that estimate $f(x) \pm 2\sigma(x)$ boundaries, obtained with the proposed method.}
    \end{subfigure}
    
    \caption{Different types of confidence analysis on a toy heteroscedastic dataset \emph{(a)}. Uniform PI \emph{(b)} estimate per-sample variance of outputs incorrectly, while confidence intervals \emph{(c)} estimate variance of model predictions that is different from the variance of outputs. Only the heteroscedastic prediction intervals \emph{(d)} provide a precise description of the dataset outputs distribution. ELM model predictions are used in \emph{(b-d)}.}
    \label{fig:pi_vs_ci}
\end{figure*}

The relation between the heteroscedastic prediction intervals and other methods is illustrated on Figure~\ref{fig:pi_vs_ci}.

\subsection{Prediction Intervals for Extreme Learning Machines}

The idea of this paper is to estimate the variance of heteroscedastic outputs $\sigma^2(\mathbf x)$ using a second ELM model. The model predictions $\hat{y}$ are computed by the first ELM, then the squared residuals $r^2 = (\hat{y} - y)^2$ are used as training outputs for the second ELM that learns to predict the conditional variance of outputs.

However, ELM predictions can be inaccurate, and their quality must be taken into account. For that reason, variances of the predictions for the first ELM $\sigma^2_y(\mathbf x)$ and the second ELM $\sigma^2_r(\mathbf x)$ are added to the predicted squared residuals $\hat{r}^2(\x)$ to bound the true variance of the outputs $\sigma^2(\x)$:
\begin{equation}
    \sigma^2(\x) \le \hat{r}^2(\mathbf x) + \sigma^2_r(\x) + \sigma^2_y(\x)
\end{equation}

In addition to directly estimating the input-dependent variance $\sigma^2(\x)$, this expression has the desired properties of giving larger variance for models with insufficient amount of training data. With an excessive amount of training data $\{\x_i, y_i\}, \ i \in [1,N]$, variances of the predicted residuals $\sigma^2_r(x)$ and the predicted outputs $\sigma^2_y(x)$ decrease to zero and the variance of true outputs is given by its ELM estimation: $\lim_{N \rightarrow \inf} \big( \hat{r}^2 \big) = \sigma^2(\x)$. A similar approach to the prediction intervals exist in feed-forward neural networks~\cite{nix_learning_1995}, however it is valid only for the case $N \rightarrow \inf$.

The output layer of ELM is a linear regression. Bienaym\'e formula~\cite{loeve_probability_1955,johnson_applied_2002} provides the variance of outputs in linear regression, and in ELM:
\begin{equation}
    \sigma^2_y(\mathbf x_i) = \h_i \Sigma_{\b} \h_i^T, \ i \in [1, \ N],
\end{equation}
where $\h_i$ is the hidden layer output of an ELM for an input sample $\x_i$. 

There is plethora of methods for estimating covariance $\Sigma_{\b}$ of normally distributed linear system weights $\b \sim \mathcal{N}(\hat{\b}, \Sigma_{\b})$.
The method of choice is weighted Jackknife estimator~\cite{wu_jackknife_1986}. It is unbiased, robust against heteroscedastic noise~\cite{shao_heteroscedasticity-robustness_1987,horn_robust_1998,flachaire_bootstrapping_2005,davidson_wild_2008}, as fast as an ELM, and scales well with the data size. Another good method for variance estimation is Wild Bootstrap~\cite{davidson_wild_2008} with nice theoretical properties, but it is slower as the bootstrap part requires several repetitions to converge.

\subsection{Weighted Jackknife for Big Data}

A summary of the Weighted Jackknife methods is presented below. Its inputs are an ELM hidden layer outputs $\H$ and residuals $\mathbf{r} = \y - \hat{\y}$.

\begin{eqnarray}
\mathbf{P} &=& (\H^T\H + \gamma \mathbf{I})^{-1} \label{eq:P} \\
\mathbf{S} &=& \H\mathbf{P}\\
\mathbf{H}'_i &=& \frac{\mathbf{r}_i^2}{1 - \mathbf{S}_i \cdot \H_i} \H_i, \ i \in [1,N] \label{eq:w}\\
\mathbf{A} &=& \H'^T\mathbf{S}\\
\Sigma &=& \mathbf{P}\mathbf{A}
\end{eqnarray}

The method uses three auxiliary matrices: $\mathbf{P}$, $\mathbf{S}$ and $\mathbf{A}$. Equation~\eqref{eq:w} creates a weighted data matrix $\H'$ by scaling every row of the original data $\H$, its denominator includes a dot product between two vectors $\mathbf{S}_i \cdot \H_i$. 

Weighted Jackknife works well together with ELM and Big Data. First, an auxiliary matrix $\mathbf{P}$ in~\eqref{eq:P} is an inverse of the already computed matrix in an ELM solution~\eqref{eq:ELM1}. 

Second, Big Data applications with huge number of samples are often limited by memory size, especially if the matrix computations are run on GPUs with very limited memory pool. Weighted Jackknife avoids such limitation by batch computations. Let the data matrix $\H$ split in $k$ equal parts with $N/k$ samples each:
\begin{equation*}
\H = \begin{pmatrix}
  \H^1 \\
  \H^2 \\
  \cdots \\
  \H^k
 \end{pmatrix}
\end{equation*}
Then auxiliary matrix $\mathbf{S}$ can be computed in the corresponding parts $\mathbf{S}^j = \H^j\mathbf{P}, \ j \in [1,k]$, and an auxiliary matrix $\mathbf{A}$ becomes a summation over all the parts $\mathbf{A} = \sum_{j=1}^k (\H'^{j})^T\mathbf{S}^j$. Size of matrices $\mathbf{A}$ and $\mathbf{P}$ does not depend on the number of samples $N$, and the weighting~\eqref{eq:w} may be done in-place without consuming additional memory. 

Having only one data part $\H^j$ in memory at a time reduces the total memory requirements by a factor of $k$. Large enough $k$ allows a single workstation to process billions of samples with Weighted Jackknife, the same way as presented for ELM in~\cite{akusok_highperformance_2015}. The practical value of $k$ is limited by the minimum size $N/k$ of a single batch, that cannot fully utilize CPU/GPU computational potential for small data batches of $N/k < 1000$~\cite{akusok_highperformance_2015}.

\subsection{ELM Prediction Intervals Algorithm}
\label{sec:summary}

Prediction intervals are computed in two stages. The first stage uses training data to learn the two necessary ELM models $m_\text{data}, \ m_\text{var}$, and estimate the covariances of output weights $\Sigma_y, \ \Sigma_r$ in these models: 

\begin{enumerate}
    \item Train an ELM model $m_\text{data}$ on the training data $\X, \y$
    \item Predict outputs $\hat{\y}$ for the training data
    \item Use weighed Jackknife to estimate covariance $\Sigma_y$ of the output weights $\b_\text{data}$
    \item Compute residuals $\r = \y - \hat{\y}$ for the training data
    \item Train another ELM model $m_\text{var}$ to predict the residuals $\X, \r^2$ \item Use weighed Jackknife to estimate covariance $\Sigma_r$ of the output weights $\b_\text{var}$
\end{enumerate}
The training data $\X, \y$ and auxiliary vectors $\hat{\y}, \r, \hat{\r}^2$ can be discarded at this point.

The second stage uses the previously trained models to predicts test outputs, their squared residuals and all variances. Then the prediction intervals are estimated with an equation~\eqref{eq:4}.

\begin{enumerate}
    \item Compute the hidden layer outputs $\H_{\text{data}}, \H_{\text{var}}$ for test inputs $\X_\text{test}$ \\ using the two ELM models $m_\text{data}, \ m_\text{var}$
    
    \item Predict test outputs $\hat{\y}_\text{test} = \H_{\text{data}} \b_\text{data}$
    \item Compute variance of the predicted outputs $\s^2_y = \text{diag} (\H_{\text{data}} \Sigma_\text{data} \H_{\text{data}}^T)$

    \item Predict squared residuals $\hat{\r}^2 = \H_{\text{var}} \b_\text{var}$
    \item Compute variance of the predicted square residuals $\s^2_r = \text{diag} (\H_{\text{var}} \Sigma_\text{var} \H_{\text{var}}^T)$
    
    \item Compute prediction intervals for a desired confidence level $\alpha$: \begin{equation}
        \text{PI} = \hat{\y}_\text{test} \pm \Phi^{-1}(\alpha) \sqrt{ \hat{\r}^2 + \s^2_r + \s^2_y }
        \label{eq:8}
    \end{equation}
\end{enumerate}

Models $m_\text{data}, \ m_\text{var}$ can have different optimal number of neurons, that should be validated. Using L2 regularization prevents numerical instabilities. Note that the predicted squared residuals $\hat{\r}^2$ might have negative values, that are replaced by zero.

\section{Experimental Results}
\label{sec:experimental}

\subsection{Artificial Dataset}

\begin{figure}
    \centering
    \includegraphics[width=0.6\textwidth]{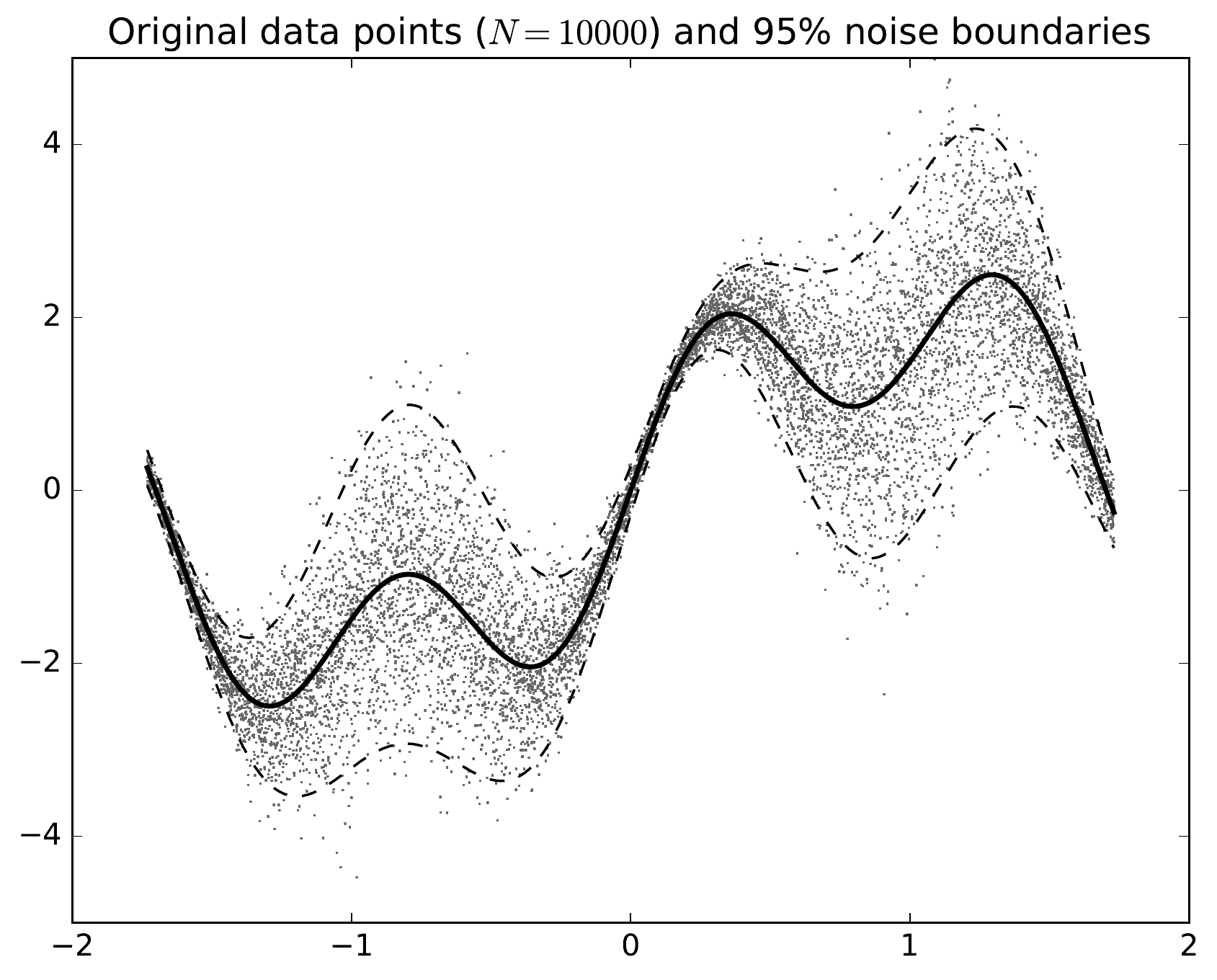}
    \caption{Artificial dataset with true 95\% intervals for noise.}
    \label{fig:mydata}
\end{figure}

An artificial dataset with heteroscedastic noise is shown on Figure~\ref{fig:mydata}. Additional tests are done on homoscedastic versions of the same dataset with the same projection function $f()$ with an input-independent normally distributed noise. All experiments used ELM with one linear and 10 hyperbolic tangent hidden neurons, in both $m_\text{data}$ and $m_\text{var}$.

\begin{figure}
    \centering
    \includegraphics[width=0.49\textwidth]{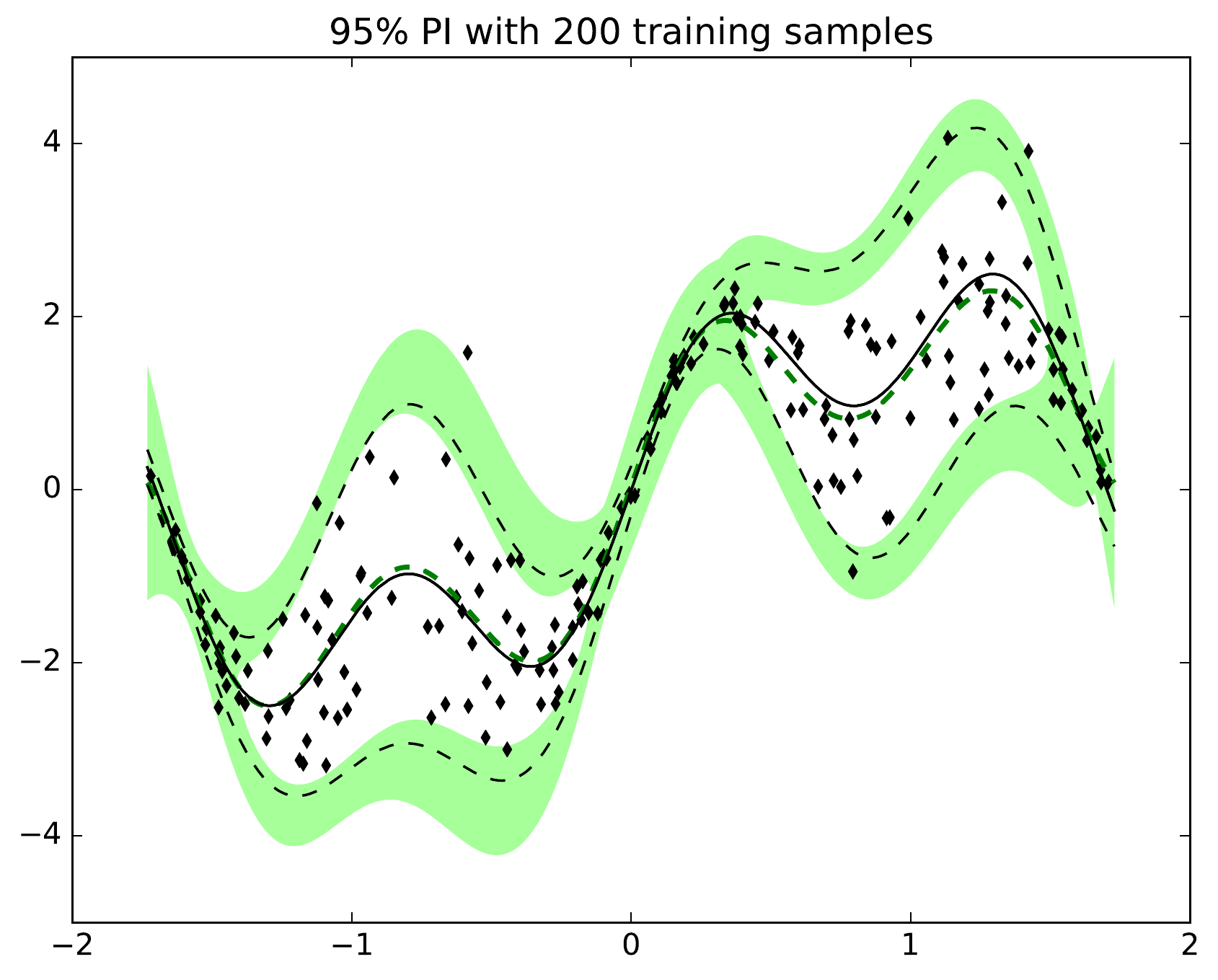}
    \includegraphics[width=0.49\textwidth]{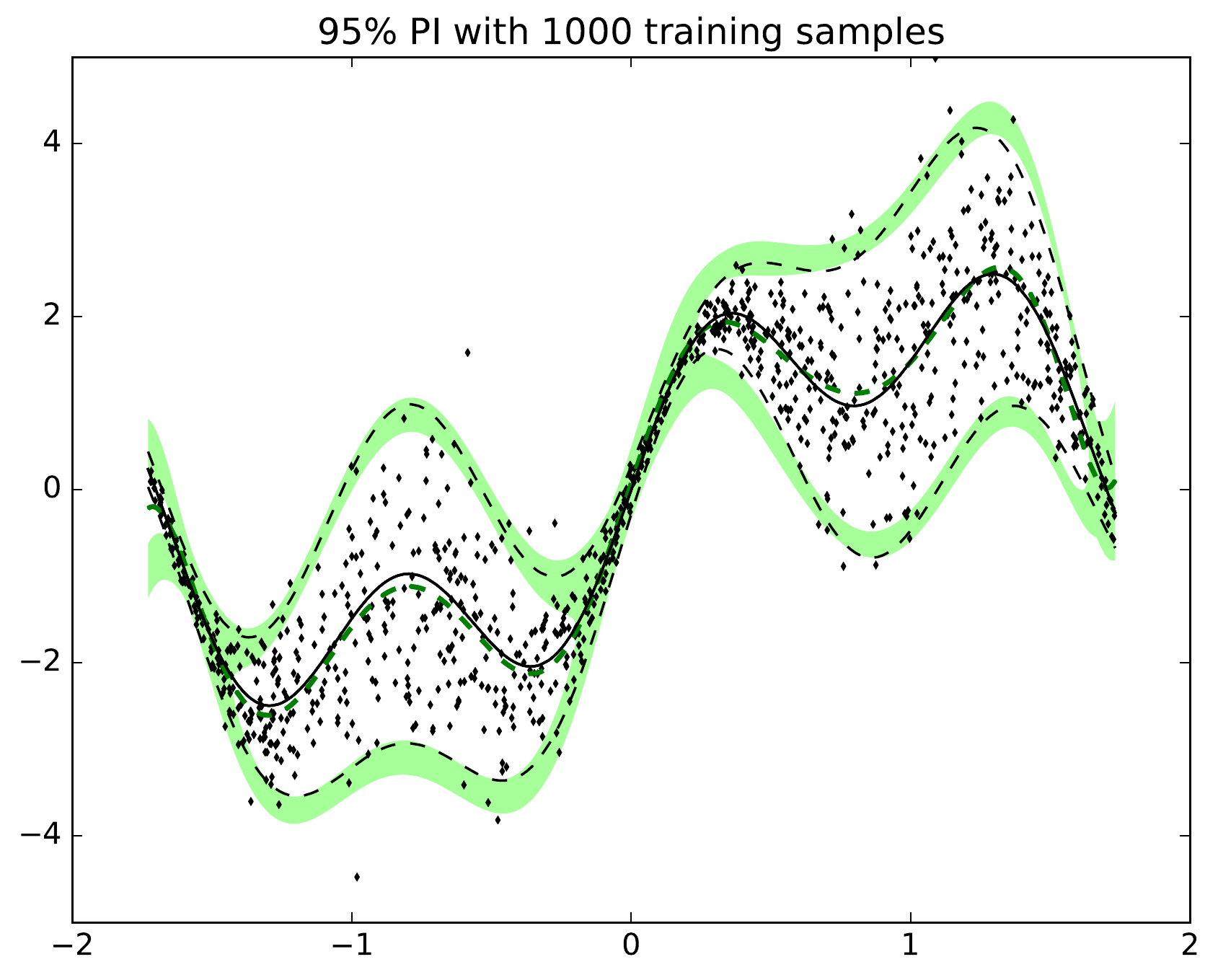}
    \caption{Estimated PI for heteroscedastic stochastic outputs. Variance of the predicted residuals $\hat{\r}^2$ (shaded area) captures model uncertainty with less training data. Thin dash lines are actual PI, solid line is the projection function, thick dash line is an estimated output, and black dots are training data samples.}
    \label{fig:myresults}
\end{figure}

Figure~\ref{fig:myresults} shows the computed PI on the heteroscedastic artificial dataset at 95\% confidence level. The figure also presents the standard deviation of the predicted residuals $1.96\sigma^2_r$ at 95\% confidence, to show how it is affected by the amount of training data. As the amount of training data increases, PI are given more precisely by $\hat{\r}^2$ and depend less on $\sigma^2_r$ (Figure~\ref{fig:myresults}, right).

\begin{figure}
    \centering
    \includegraphics[width=0.49\textwidth]{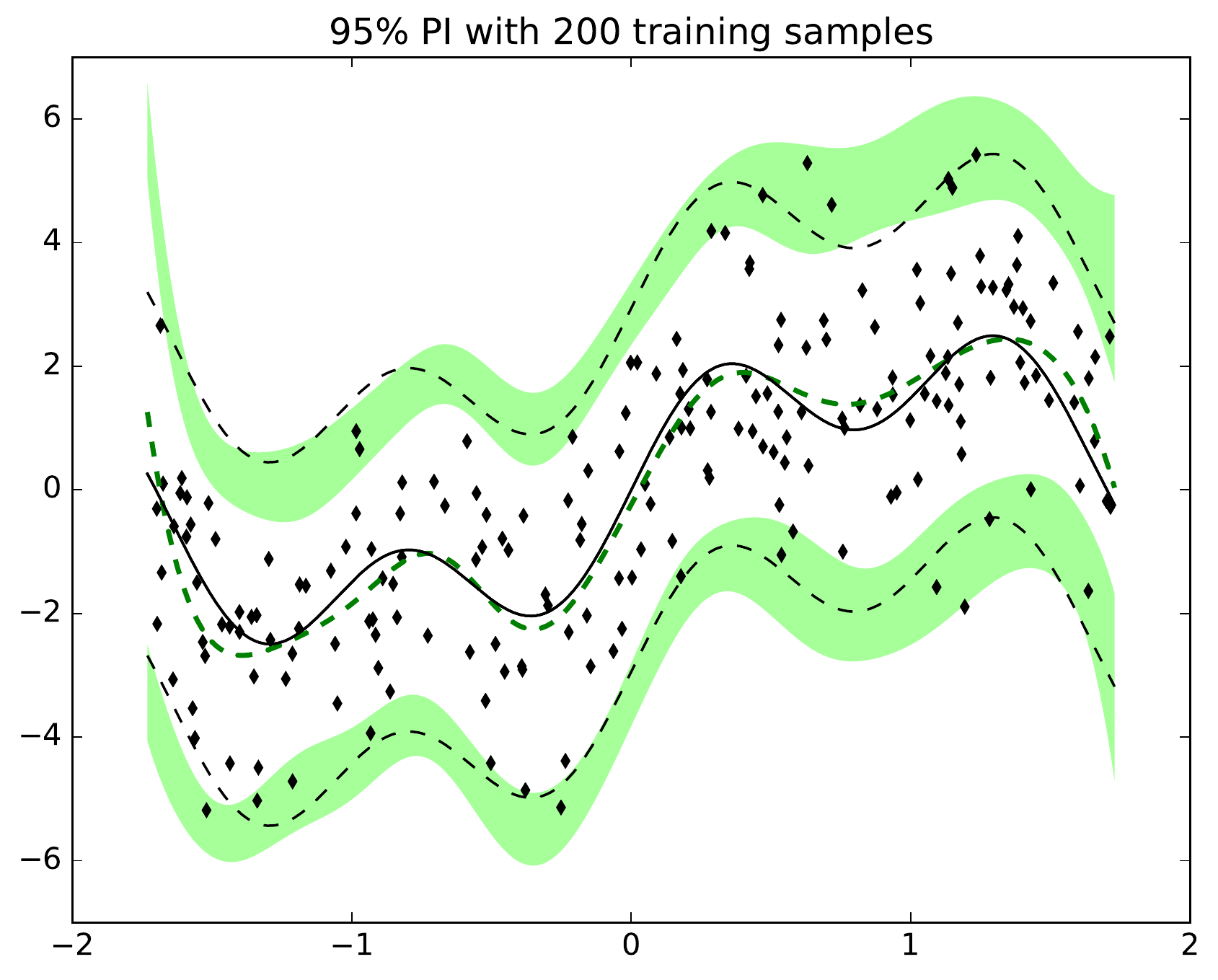}
    \includegraphics[width=0.49\textwidth]{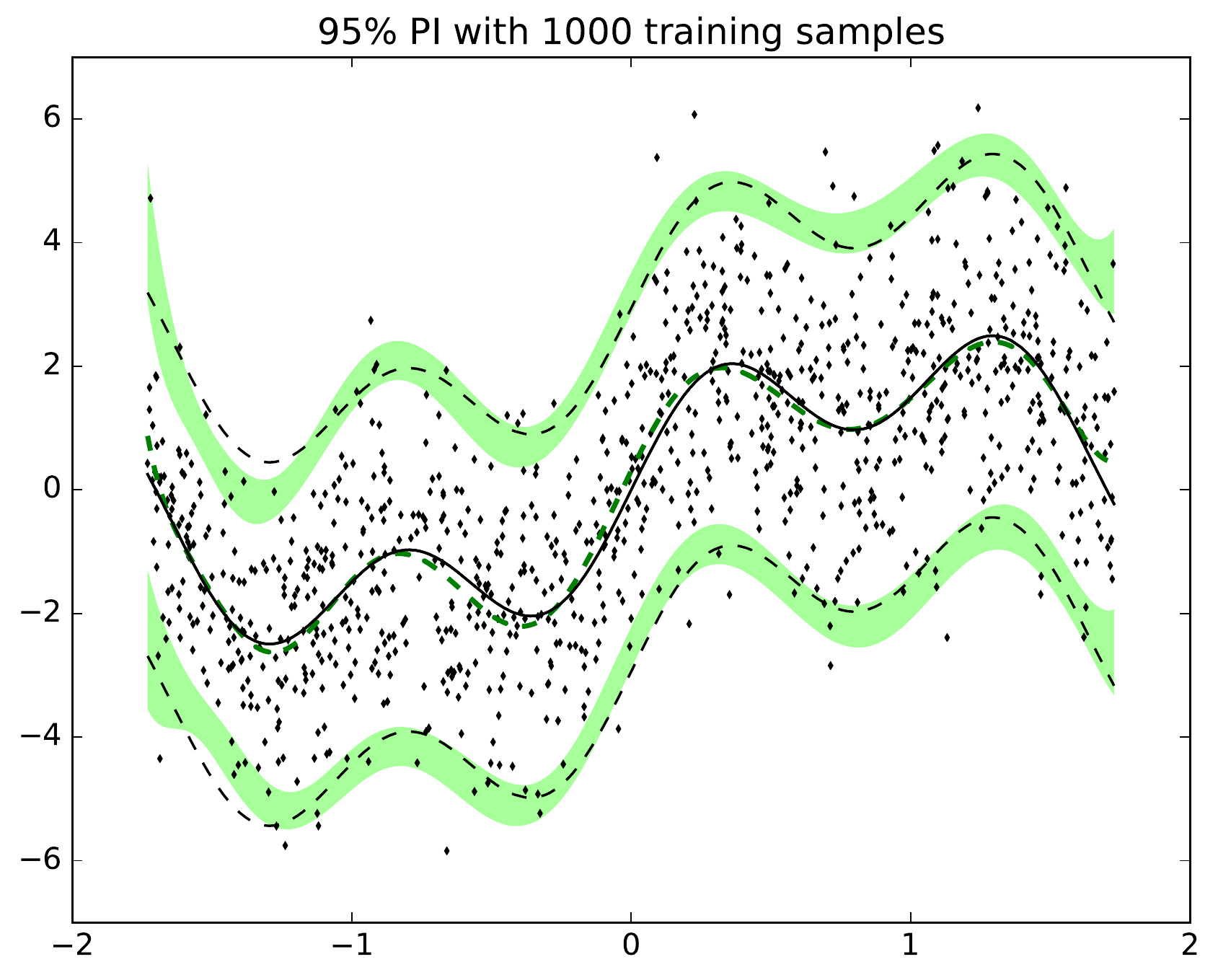}\\
    \vspace{10pt}
    \includegraphics[width=0.49\textwidth]{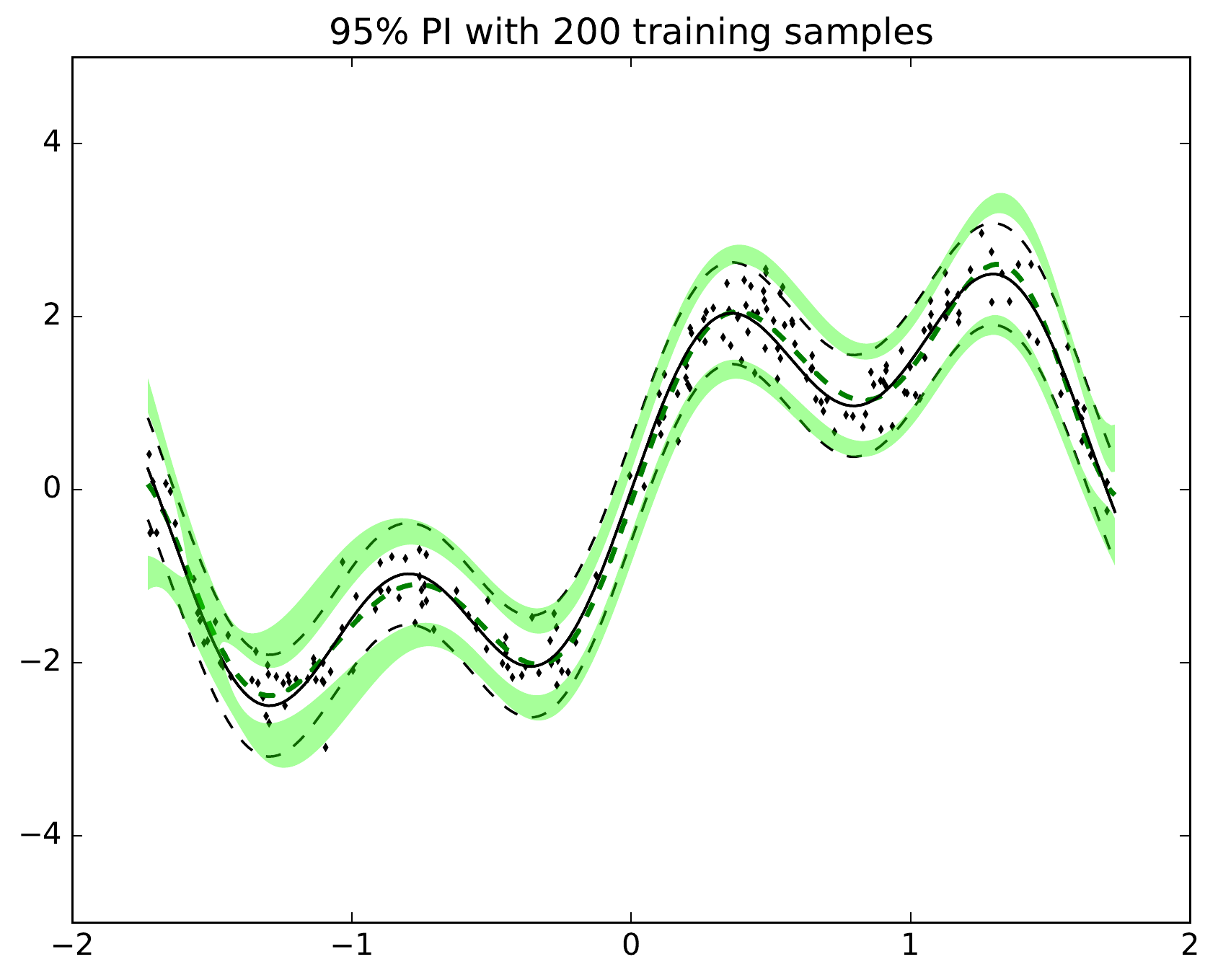}
    \includegraphics[width=0.49\textwidth]{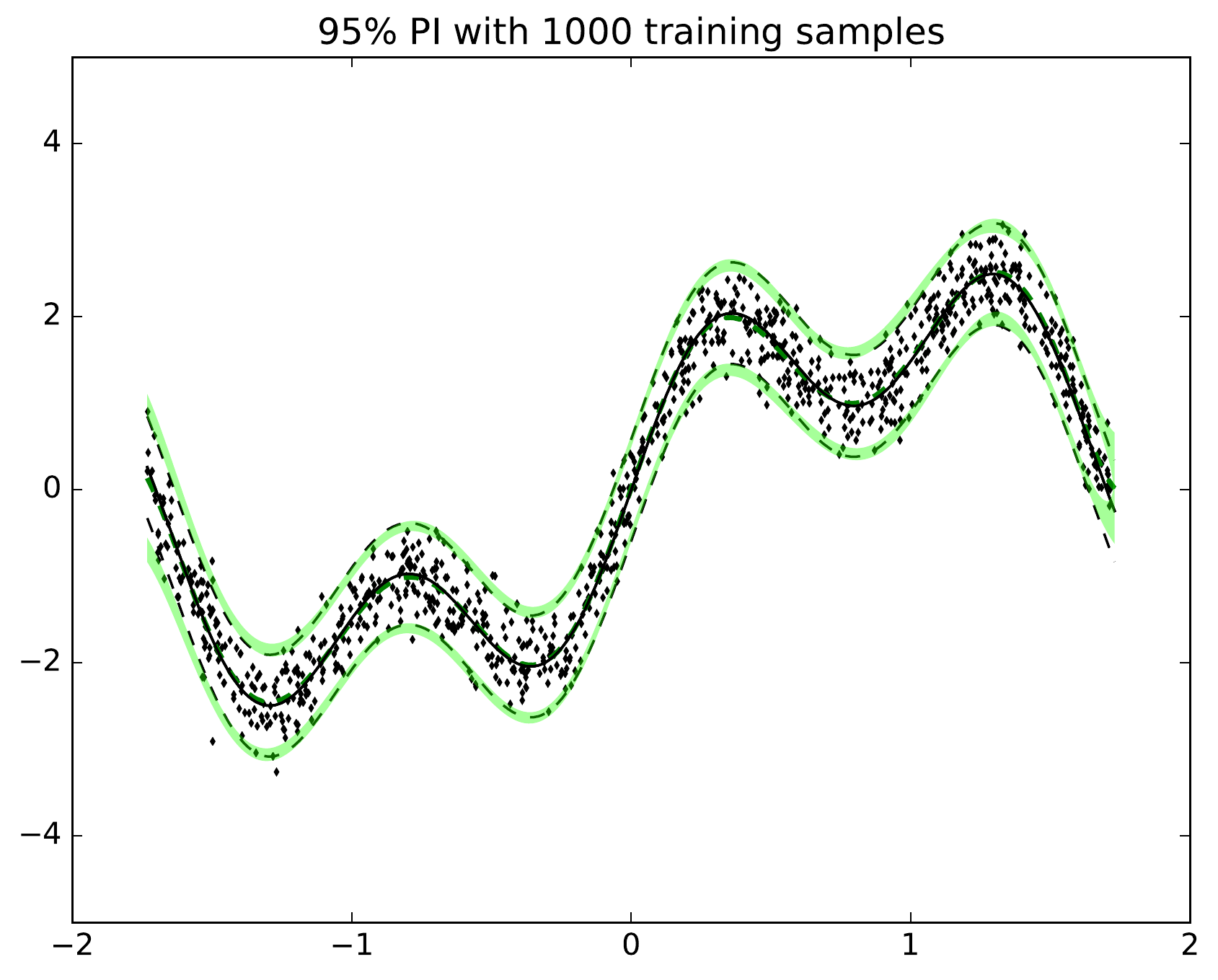}
    \caption{Estimated PI and their variance (shaded area) for homoscedastic stochastic outputs with difference variance; more data leads to more precise PI. Thin dash lines are actual PI, solid line is the projection function, thick dash line is estimated projection function, and black dots are training data samples.}
    \label{fig:myresults_homo}
\end{figure}

Similar results obtained for the datasets with homoscedastic noise, presented on Figure~\ref{fig:myresults_homo}. Larger variance of outputs makes the prediction task harder, leading to larger errors in $\hat{y}$ (Figure~\ref{fig:myresults_homo}, upper left). At the same time the variance of $\hat{\r}^2$ increases (Figure~\ref{fig:myresults_homo}, shaded area), and the true PI rarely go beyond their estimated boundaries. Smaller variance of noise leads to more more precise PI, that still cover the true PI most of the time.

\begin{figure}
    \centering
    \includegraphics[width=0.49\textwidth]{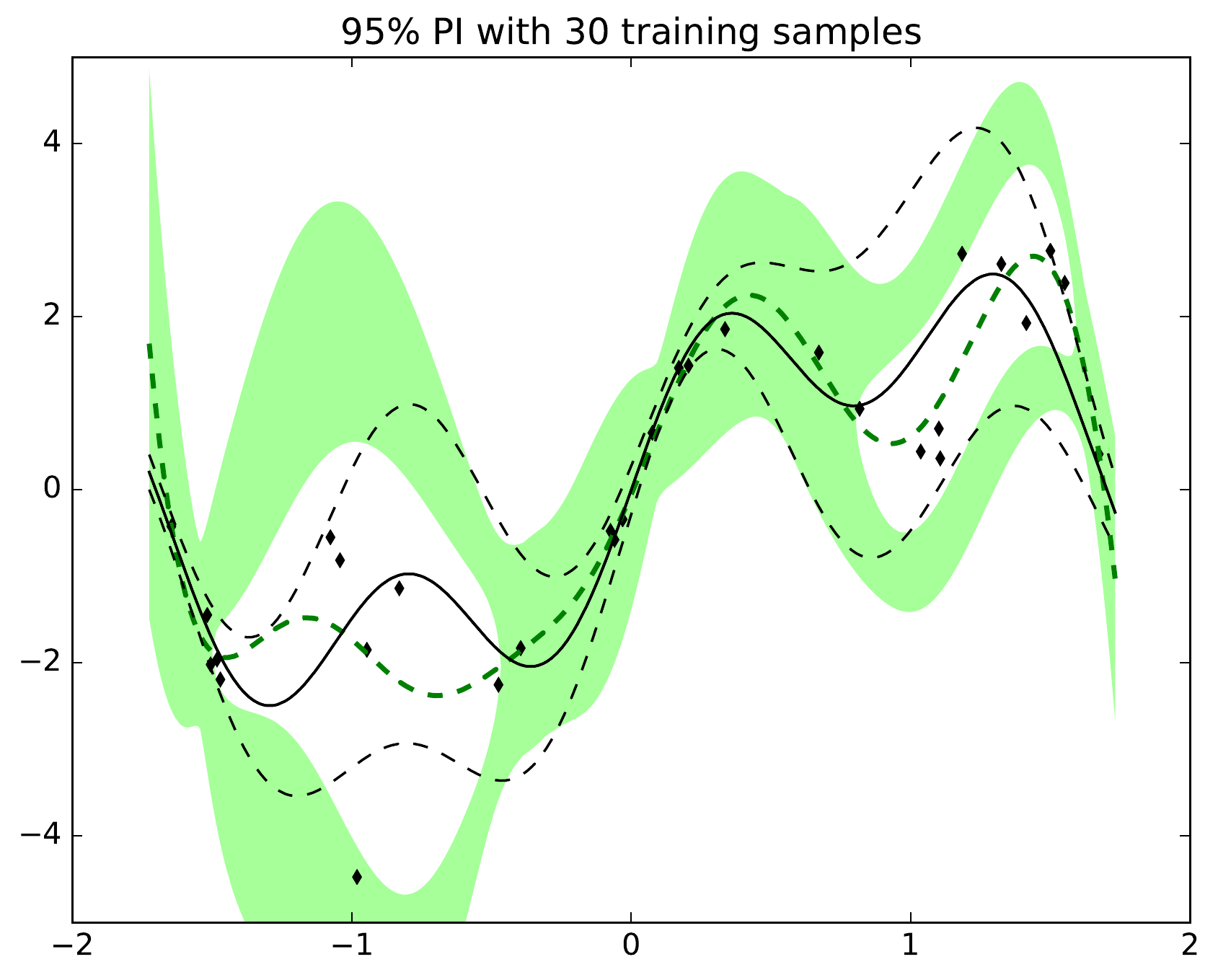}
    \includegraphics[width=0.49\textwidth]{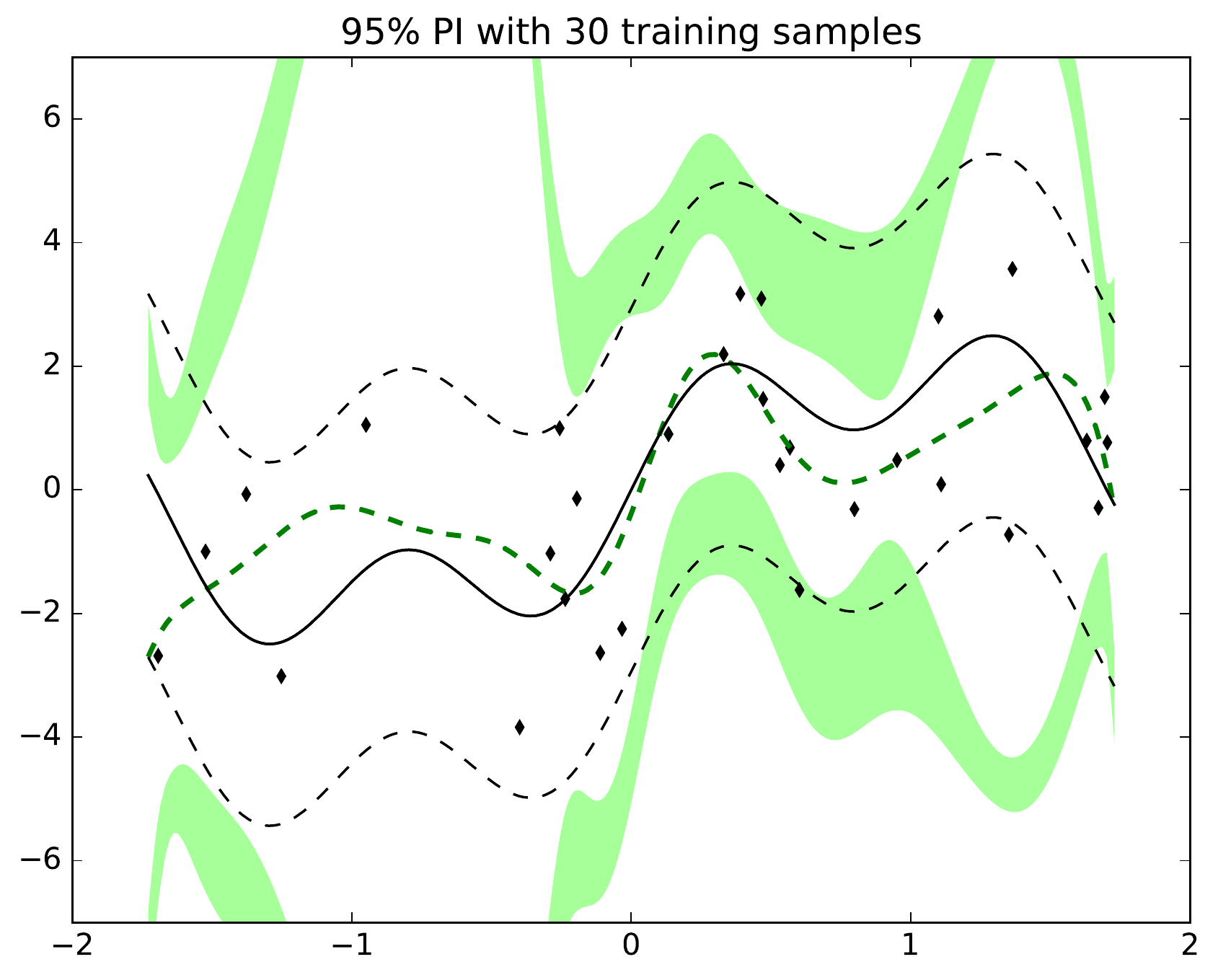}
    \caption{Estimated PI and their variance (shaded area) with an insufficient amount of training data; PI are over-estimated in poorly predicted areas. Thin dash lines are actual PI, solid line is the projection function, thick dash line is estimated projection function, and black dots are training data samples.}
    \label{fig:myresults_extreme}
\end{figure}

In the extreme case of a training set with only 30 samples (which is not enough for learning the correct shape of the true projection function), the predicted squared residuals $\hat{\r}^2$ become unreliable. However, including their variance in the predictions compensates for the model uncertainty (see Figure~\ref{fig:myresults_extreme}). It sometimes leads to over-estimation of the true PI, but this is a desired property that prevents an uncertain model from predicting false highly confident outputs $\hat{y}$.

\subsection{Comparison on Real World Datasets}

ELM Prediction Intervals are compared on four real datasets with four other methods presented in~\cite{khosravi_lower_2011}. Details of the datasets are given in Table~\ref{tab:data}. The paper uses two common metrics: Prediction Intervals Coverage Probability (PICP) that is a percentage of test samples whose outputs lie between the PI, and the Normalized Mean Predicted Interval Width that is an average width of PI on a test set divided by the range of the test targets. PICP shows what percentage of targets actually lie within PI, and it should correspond to the target coverage. NMPIW presents how optimal are the PI for the given task, compared to a naive approach of simply taking the full range of targets as an interval. Ideal PI have a small NMPIW with PICP equals to $\Phi^{-1}(\alpha)$ target coverage.

\begin{table}
\centering
\caption{Real-world datasets used for comparison.}
\label{tab:data}
\begin{tabular}{l c c c}
    \hline
    Dataset     & Samples         & Features & Reference \\
    \hline
    Concrete compressive strength & 1030 & 8 & \cite{yeh_modeling_1998} \\
    Plasma beta-carotene  & 315   & 12 & \cite{nierenberg_determinants_1989}\\
    Powerplant - Steam pressure  & 200  & 5 & \cite{guidorzi_identification_1974} \\
    Powerplant - Main steam temperature & 200 & 5 & \cite{guidorzi_identification_1974}  \\
    \hline
\end{tabular}
\end{table}

The two measures PICP and NMPIW are inter-dependent as increasing PI width also increases the coverage. The comparison work~\cite{khosravi_lower_2011} proposed a combined measure to replace PICP and NMPIW, but it is subjective due to two arbitrary hyper-parameters. This paper rather presents PICP and NMPIW on the same plot.

ELM PI method proposed in the paper is compared to four other methods of computing PI for neural networks. The Delta method~\cite{chryssolouris_confidence_1996} linearizes a neural networks model around a set of parameters, then applies an asymptotic theory to construct the PI. An extension of the Delta method to heteroscedastic noise is available~\cite{ding_backpropagation_2003}, although still limited due to linearization. Bayesian learning of neural network weights allows for direct derivation of variance for particular predicted values~\cite{mackay_evidence_1992}, but at a very high computational cost. Bootstrap method is directly applicable to any machine learning method including neural networks, although caution should be taken in selecting bootstrap parameters to make the method resilient to heteroscedastic noise~\cite{davidson_wild_2008}. Finally, the Lower Upper Bound Estimation (LUBE) method proposed by~\cite{khosravi_lower_2011} uses two additional outputs in a neural network to predict lower and upper PI, training the network with a custom cost function that includes both PICP and NMPIW.

Experimental setup uses L1 regularized ELM model~\cite{miche_opelm_2010} for automatic model structure selection on relatively small datasets, implemented in HP-ELM toolbox~\cite{akusok_highperformance_2015}. The datasets are randomly split in 70\% training and 30\% test samples, median results over 30 initializations are reported. Numerical experimental results are given in Table~\ref{tab:results}; comparison numbers for other methods are available in the corresponding paper~\cite{khosravi_lower_2011}. Runtime is reported for a 1.4GHz dual-core laptop.

\begin{table}
\centering
\caption{Experimental results of ELM Prediction Intervals.}
\label{tab:results}
\begin{tabular}{l c c c}
    \hline
    Dataset  & PICP(\%) & NMPIW(\%) & Runtime(ms) \\
    \hline
    Concrete compressive strength & 91.59 & 34.01 & 92 \\
    Plasma beta-carotene  & 92.63   & 40.66 & 36 \\
    Powerplant - Steam pressure  & 93.33  & 39.29 & 27 \\
    Powerplant - Main steam temperature & 88.33 & 18.38 & 35 \\
    \hline
\end{tabular}
\end{table}

\begin{figure}
    \centering
    \includegraphics[width=0.49\textwidth]{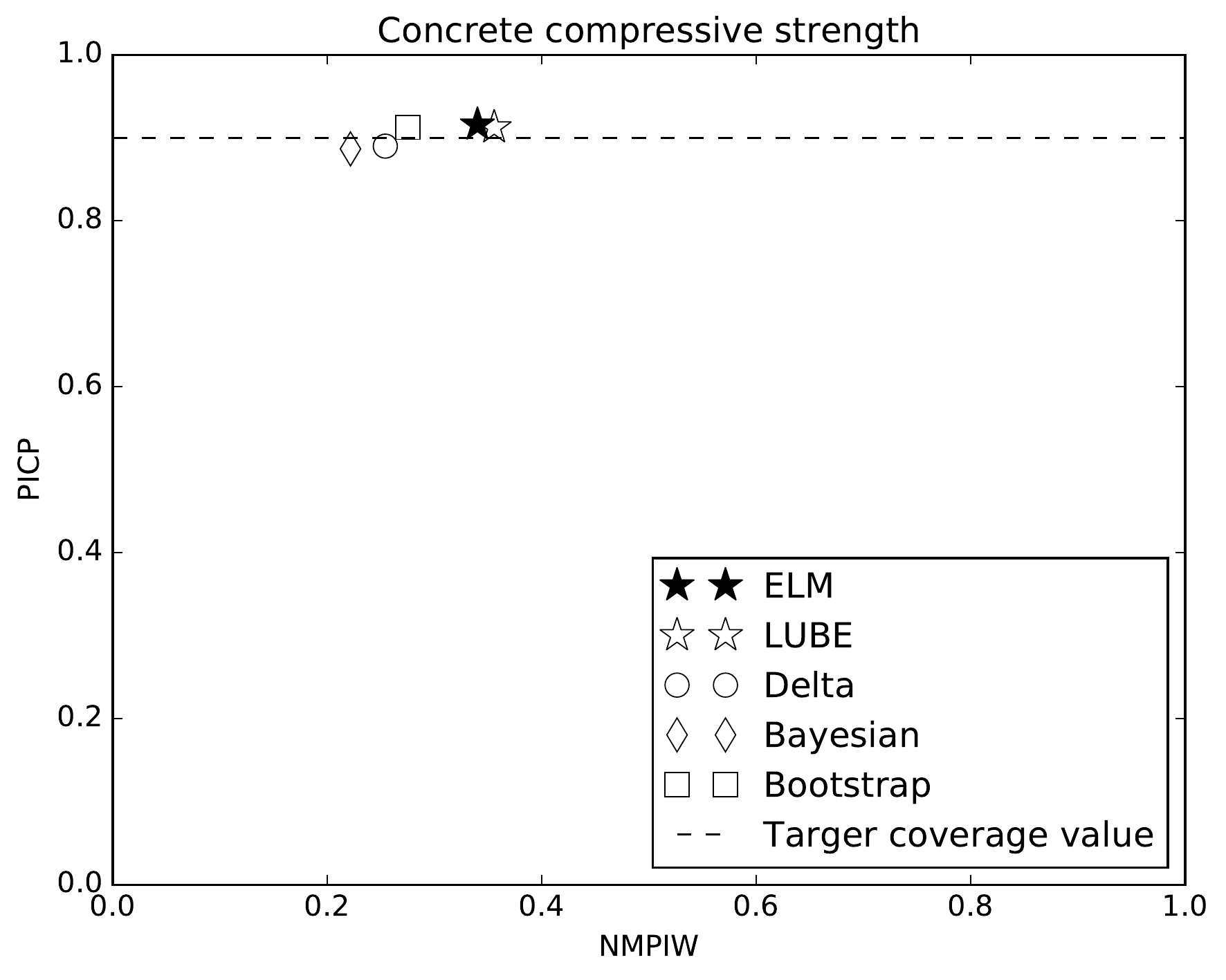}
    \includegraphics[width=0.49\textwidth]{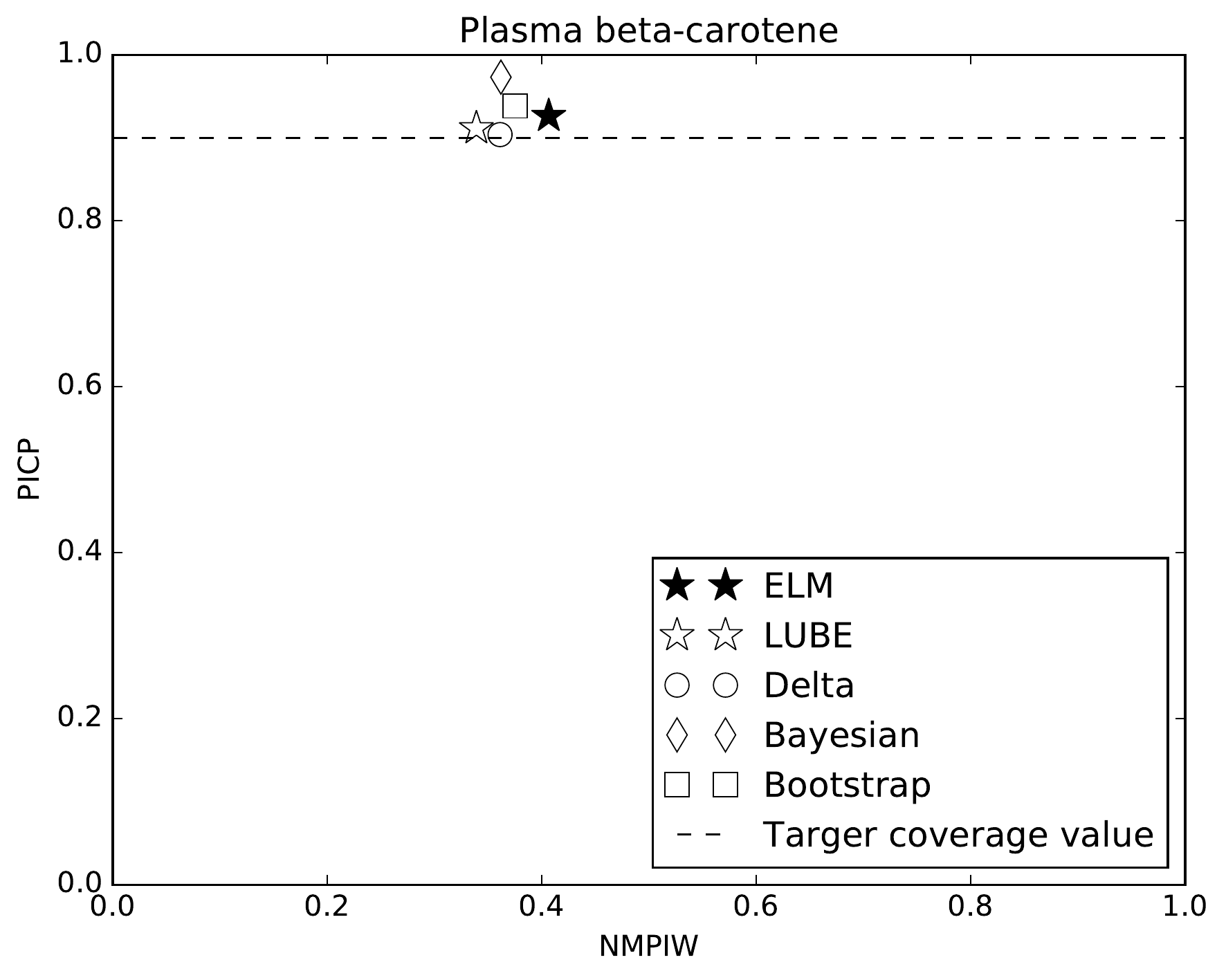}\\
    \includegraphics[width=0.49\textwidth]{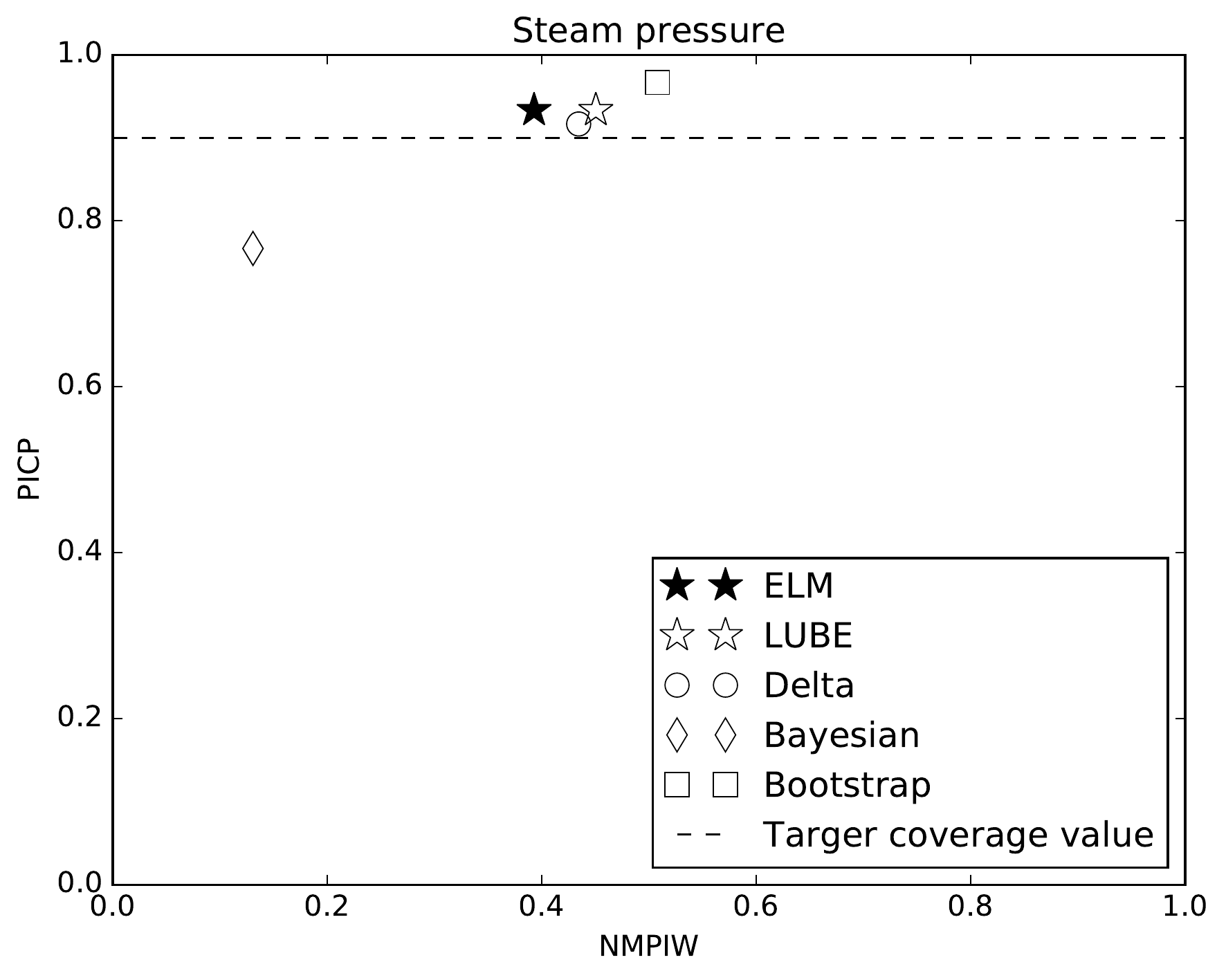}
    \includegraphics[width=0.49\textwidth]{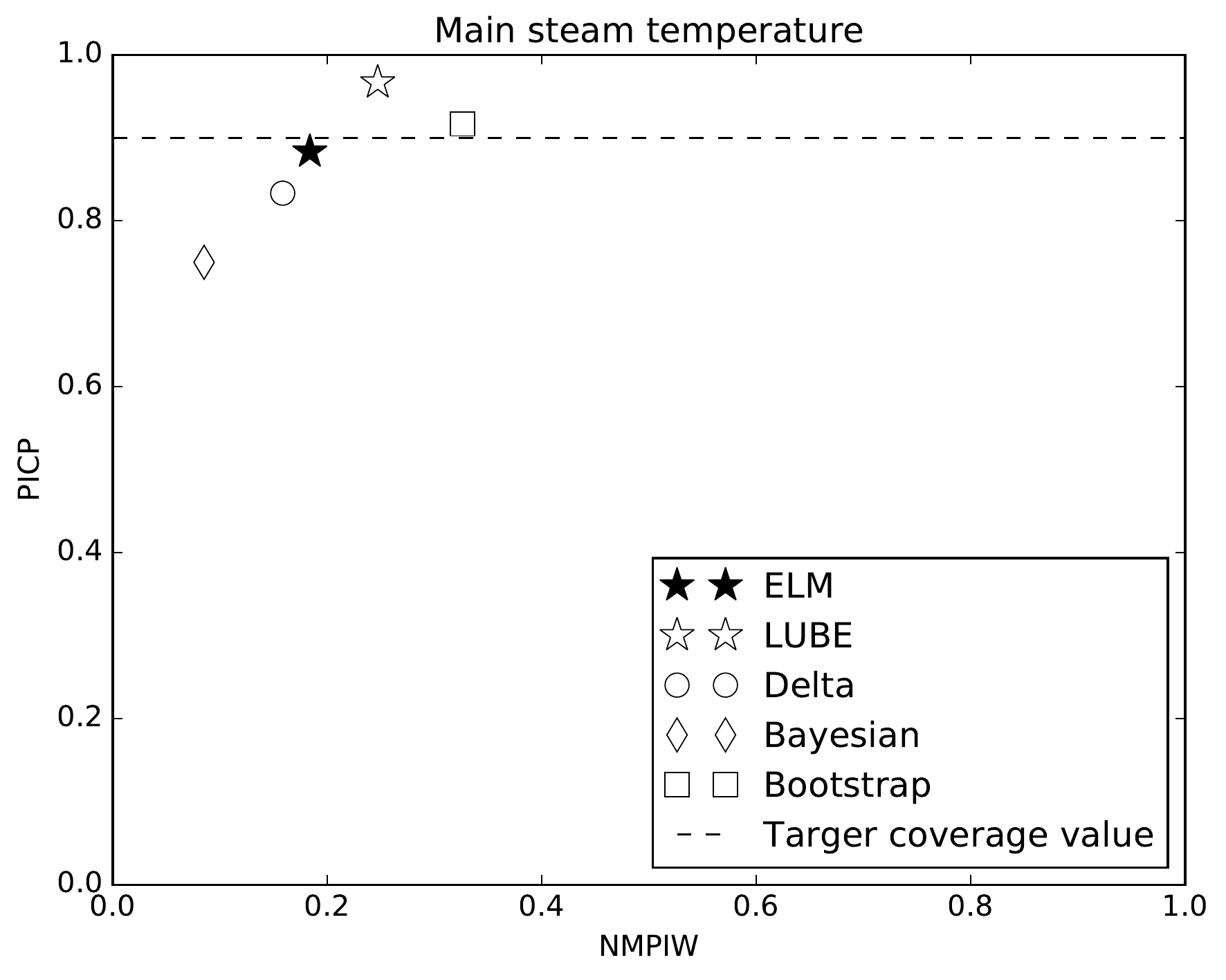}
    \caption{Comparison of the ELM PI method (\emph{filled star}) with four other methods from~\cite{khosravi_lower_2011}. Best performing methods have low NMPIW and the target coverage (points close to the upper left corner).}
    \label{fig:real_comparison}
\end{figure}

Performance of the methods is shown as points in NMPIW/PICP coordinates, presented on Figure~\ref{fig:real_comparison}. An ideal method would be at the left edge of the dashed line (low NMPIW with precise PICP). As shown on the figure, ELM PI method performs better on Steam pressure dataset, a little worse on Plasma beta-carotene datasets, and about average on the other two.

\begin{figure}
    \centering
    \includegraphics[width=0.49\textwidth]{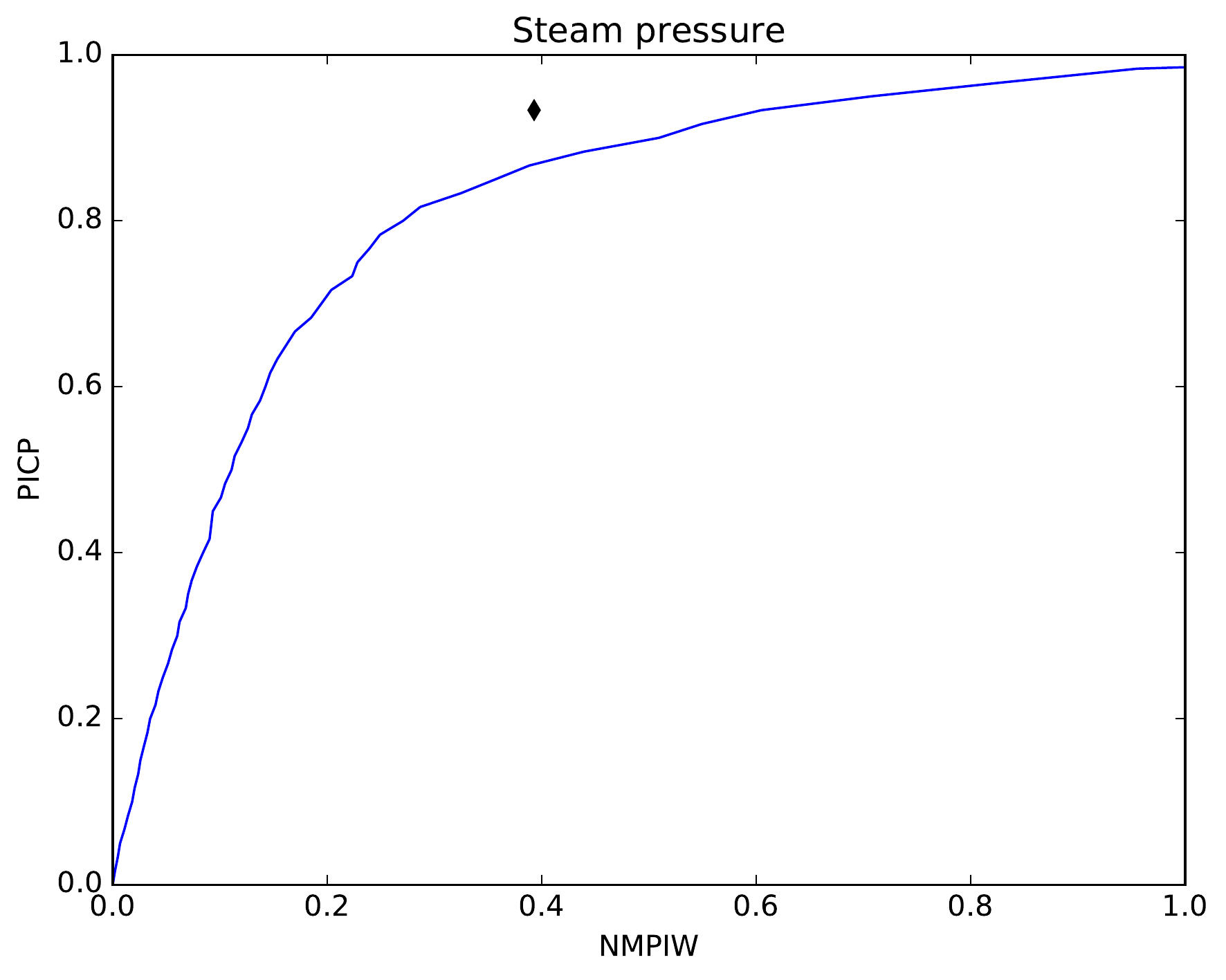}
    \includegraphics[width=0.49\textwidth]{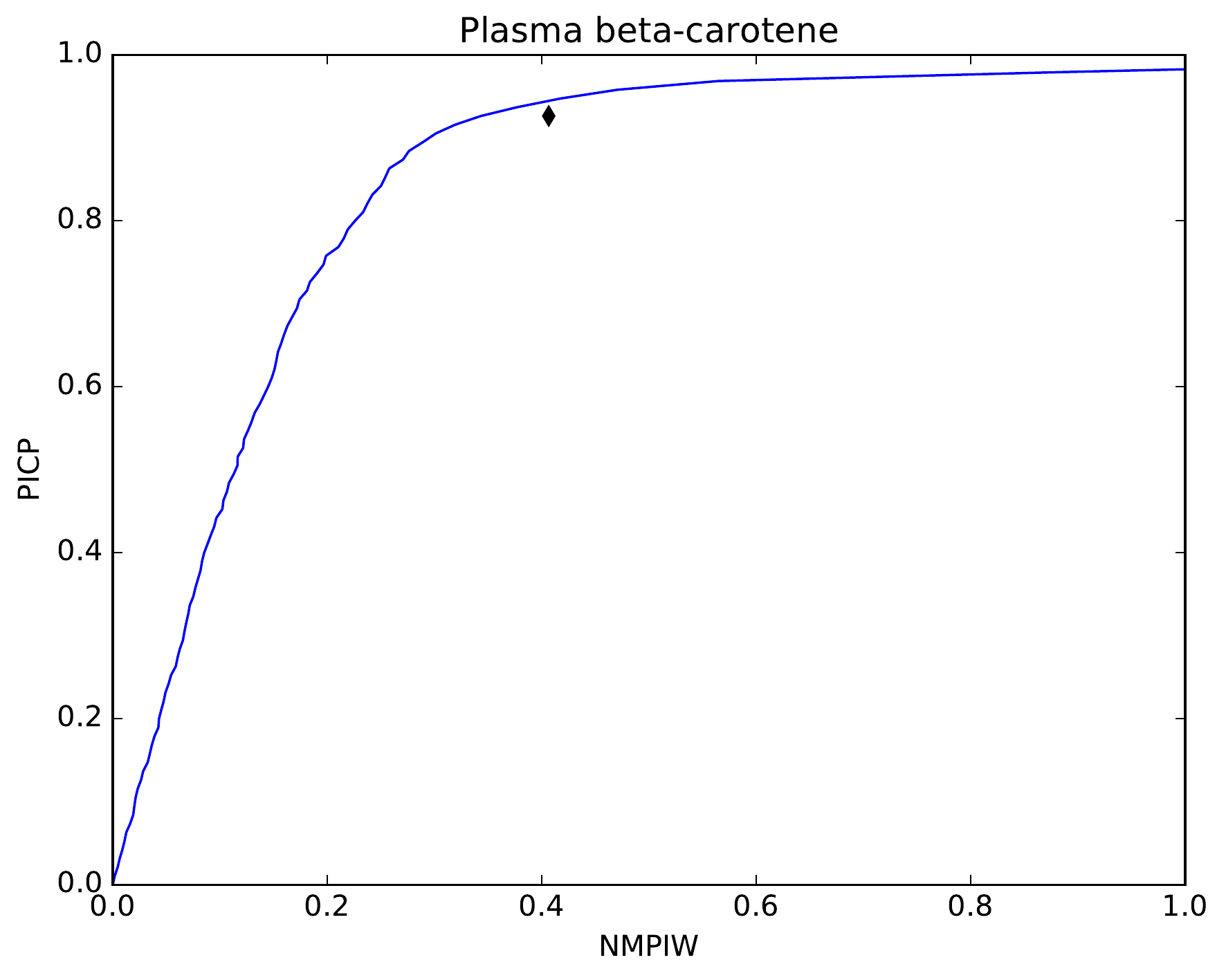}
    \caption{Comparison of ELM PI (\emph{black marker}) with uniform PI of varying width (\emph{solid line}). Heteroscedastic ELM PI perform better on the Steam pressure dataset, while uniform PI are enough for the Plasma beta-carotene dataset.}
    \label{fig:real_analysis}
\end{figure}

A further analysis shows possible reasons for good performance on Steam pressure, and bad one on Plasma beta-carotene. The analysis compares against uniform PI using the same ELM predictions for a dataset. Such PI estimate homoscedastic noise correctly, but cannot learn heteroscedastic noise. Let a uniform PI grow starting from zero, then as they grow both coverage and the interval width will increase, generating many pairs of \{NMPIW, PICP\} points. These points are then connected by a line that represents homoscedastic PI performance boundary. Homoscedastic PI performance boundary and ELM PI for the two datasets in question are shown on Figure~\ref{fig:real_analysis}.

Obviously, useful heteroscedastic PI must be above this boundary -- but in practice they may end up below due to poorer parameter estimation. Indeed, heteroscedastic PI need interval width per sample while homoscedastic PI only have interval width per dataset, that is easier to estimate precisely. As seen from Figure~\ref{fig:real_analysis}, this is the situation for ELM PI on the Plasma beta-carotene dataset where uniform PI perform better. On Steam pressure however, heteroscedastic PI perform better than uniform ones as they have higher coverage with the same average width. Another possible reason for the difference in performance is that Plasma beta-carotene dataset has homoscedastic noise, while Steam pressure dataset has actually a heteroscedastic noise (or heteroscedastic stochastic outputs), so heteroscedastic PI provide the most benefit when computed on the latter dataset.

\section{Minimizing False Positives on a Large Real Dataset}

This experiment uses PI to minimize the amount of false positive predictions on a large classification task. Note that the proposed PI methodology applies equally well to regression, and monotonic classification tasks are handled even better using purposely developed~\cite{ZHU2017205} implementations of ELM as $m_\text{data}$.

A 4,000,000-sample dataset of pixel colors for skin/non-skin classification is created from the FaceSkin Images dataset~\cite{phung_skin_2005}. The inputs are colors of the target pixel and its $7 \times 7$ neighbors with $7 \times 7 \times 3 \ \text{(RGB)} = 147$ input features total, and the outputs are +1 for skin pixels and -1 for non-skin ones. The dataset uses photos of various people under different lighting conditions, without any pre-processing. True skin masks are created manually and are highly accurate. Half of the dataset is used for training, and the other half for test.

The applied ELM model uses 147 linear + 200 sigmoid neurons. Predictions of ELM are real values, that are turned into classes by taking their sign. Due to a simple model and input features (that are not tailored for image processing) the performance is average at about 87\% accuracy. The goal of the experiment is to check whether the per-sample PI can be used to significantly improve the accuracy at a cost of coverage, compared to per-datasets PI computed by MSE.

To trade coverage for precision, a threshold $\theta$ is introduced. ELM predictions with an absolute value less than $\theta$ are ignored. A value of $\theta$ corresponding to the desired coverage percentage is found by scalar optimization methods. For per-sample PI, threshold $\theta$ is multiplied by the value of the corresponding $\text{PI}_i$ for a prediction $y_i$.

\begin{figure}
    \centering
    \includegraphics[width=\textwidth]{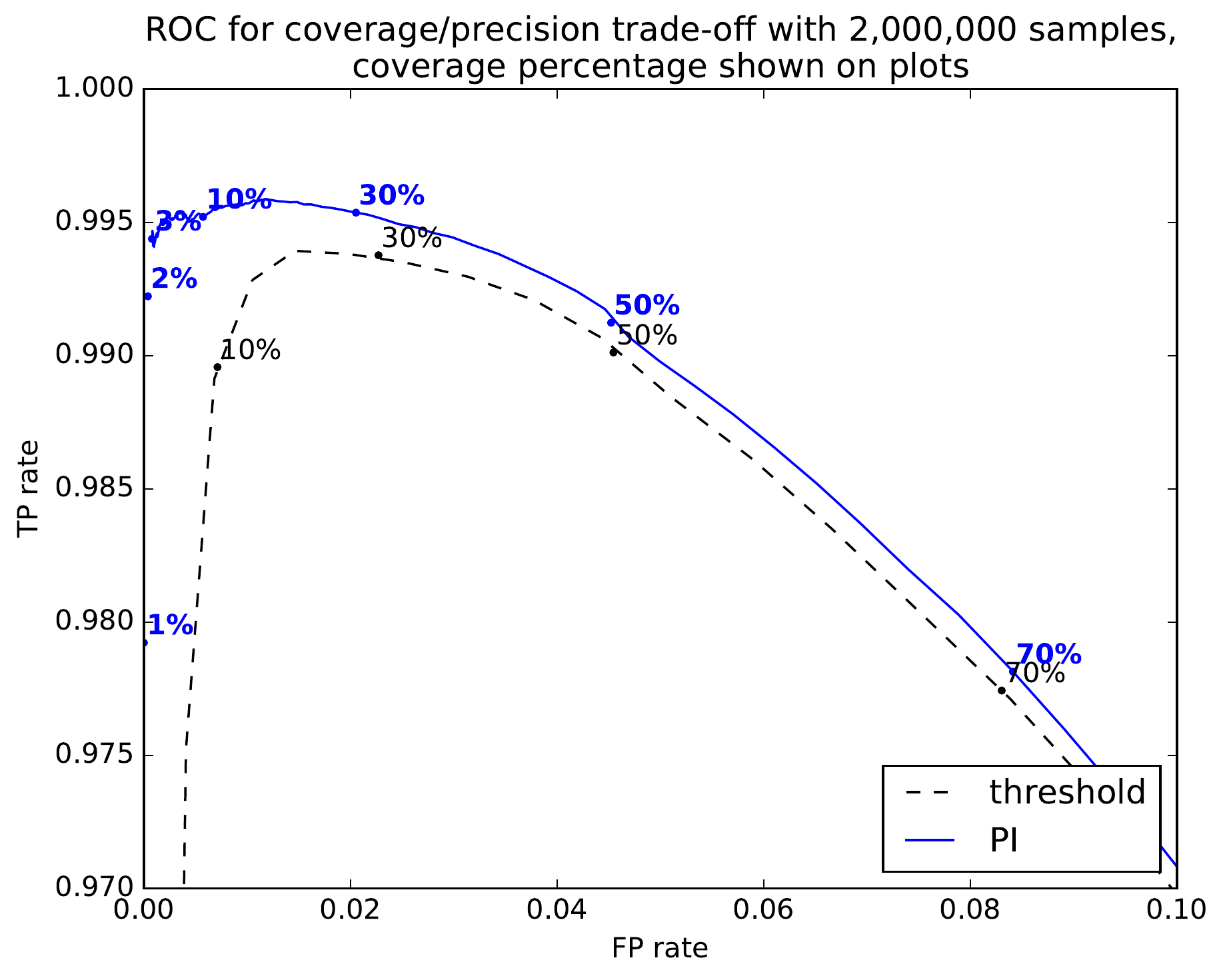}
    \caption{True Positive versus False Positive rate for the most confident part of the predictions (depicted by percentage) for a MSE-based threshold (dash line), and sample-specific threshold based on PI (solid line). Per-sample PI give almost zero False Positives for 3\% best predictions, and exactly zero for 1\% best. True Positives rate is overall higher than for an MSE-based threshold.}
    \label{fig:skin_large}
\end{figure}

The results are shown on Figure~\ref{fig:skin_large}. Here, an ELM models with a total of 347 hidden neurons is trained on a dataset with two million samples. The per-sample PI improves the true positive rate slightly. However, they reach almost zero false positives with 3\% coverage, and exactly zero at 1\%. Contrary to the proposed method, uniform PI computed with MSE cannot achieve zero false positives. Although one percent of coverage seems very little, it represents 20,000 test samples for that dataset, and it is a surprising achievement for a simple ELM model that is not optimized for False Positives reduction like in custom applications~\cite{akusok_twostage_2014}. A specifically designed model, or an ensemble of multiple models could achieve zero False Positives with a larger coverage -- a significant result for practical use of ELM, and Machine Learning algorithms in general.

\subsection{Runtime Analysis}

The runtime of per-sample PI is examined on the pixel classification dataset explained above. The experiments are run on a desktop machine with 4-core Intel Skylake CPU, using an efficient ELM toolbox from~\cite{akusok_highperformance_2015}. With 2,000,000 training samples and 347 hidden neurons, training an ELM takes 12 seconds (for both $m_{\text{data}}$ or $m_{\text{var}}$). Computing covariance matrices $\Sigma_y$ and $\Sigma_r$ with weighted Jackknife method takes 25 seconds each, or only twice longer that training an ELM itself. Test predictions take 8 seconds to compute, and test per-sample PI take 32 seconds. In total, prediction intervals increase the ELM runtime by a constant factor of about 5.

Runtime on the real-world datasets is not directly comparable with the other methods as they are run on different machines, but it is the same order of magnitude as Bootstrap, an order of magnitude faster than Delta or Bayesian methods, but also an order of magnitude slower than the LUBE method. Replacing L1 regularized ELM with standard ELM reduces the runtime to the level of LUBE method, however it degrades the results on small datasets with a few hundreds samples. Extremely large datasets that do not need regularization benefit from the faster run speed.

\section{Conclusion}
\label{sec:conclusion}

The paper proposed a method of computing per-sample prediction intervals for Extreme Learning Machines. It successfully evaluates variance of heteroscedastic stochastic outputs, using only ELM models and the weighted Jackknife method. The proposed framework works well for homoscedastic outputs, making the proposed method  applicable on a general level. ELM PI is comparable to other methods of computing PI in neural networks on small datasets, while keeping it possible to have very fast runtimes and scalability for Big Data.

On a real dataset, the method has shown to allow for a better precision and lower False Positives rate. Heteroscedastic PI performs in a similar way as uniform PI from Mean Squared Error on 50\%-70\% of dataset samples, but they make a huge difference on the most confidently predicted 1\%-10\% of samples. For these samples, the proposed PI allowed to achieve zero False Positives rate even with a basic ELM model, which is an extremely useful feature in many practical applications. The runtime is comparable to the runtime of an ELM itself that makes it feasible for large datasets of Big Data problems.

ELM PI can be easily extended to non-symmetric PI by using two ELM models in the second stage for predicting upper and lower boundaries separately. An ensemble of ELMs may increase the coverage for zero False Positives data predictions. These extensions will be examined and evaluated in future works on this topic.

\bibliographystyle{plain}
\bibliography{pi}

\end{document}